\documentclass{article} % For LaTeX2e
\usepackage[preprint]{neurips_2026}
\usepackage{times}

\usepackage{amsmath,amsfonts,bm}

\def\eqref#1{equation~\ref{#1}}
\def\1{\bm{1}}

\DeclareMathAlphabet{\mathsfit}{\encodingdefault}{\sfdefault}{m}{sl}
\SetMathAlphabet{\mathsfit}{bold}{\encodingdefault}{\sfdefault}{bx}{n}

\usepackage{hyperref}
\usepackage{url}
\usepackage{graphicx}
\usepackage{booktabs}
\usepackage{array}
\usepackage{amsmath}
\usepackage{amssymb}
\usepackage{bbm}
\usepackage{placeins}

\title{RareDx: Controlled Knowledge Integration and Graph-Grounded Policy Optimization for Rare-Disease Diagnosis}

\author{%
  \mdseries Bo Zhang\textsuperscript{1,$\dagger$}\quad
  Yuchen Wang\textsuperscript{2,$\dagger$}\quad
  Dongbai Li\textsuperscript{3}\\
  Matthew Yu Heng Wong\textsuperscript{4}\quad
  Qingkai Zeng\textsuperscript{5}\quad
  Lijun Wang\textsuperscript{6,*}\\
  Tien-Yin Wong\textsuperscript{3}\quad
  Peng Cui\textsuperscript{3,*}\quad
  Tianyu Liu\textsuperscript{3,7,*}\\[0.5em]
  \textsuperscript{1}Xi'an Jiaotong University\quad
  \textsuperscript{2}UIUC\\
  \textsuperscript{3}Tsinghua University\quad
  \textsuperscript{4}University of Cambridge\\
  \textsuperscript{5}Nankai University\quad
  \textsuperscript{6}Zhejiang University\quad
  \textsuperscript{7}Yale University
  \\[0.3em]
  {\footnotesize $\dagger$: Equal contribution. $*$: Corresponding authors.}
}

\begin{document}

\def\method{RareDx}

\maketitle

\begin{abstract}
Rare-disease diagnosis is a long-tail reasoning problem: phenotypes are incomplete, individual
disorders are sparsely documented, and relevant evidence is distributed across ontologies,
gene annotations, and biomedical text. Language models consequently favor common conditions,
miss rare candidates, or produce plausible but invalid names. We introduce \textbf{RareDx}, which
couples controlled evidence use with knowledge-graph-grounded policy optimization.
RareDx-Harness normalizes
heterogeneous records into one ranked-diagnosis task and compares direct inference, static
retrieval, adaptive tools, and structured phenotype-gene-disease reasoning over a shared
knowledge layer. The training pipeline combines Top-10 post-training with
\textbf{RareDx-KGPO}, our knowledge-graph-grounded policy optimization method. Its reward projects
predictions into a canonical disease graph and integrates curated graded relevance, ontology proximity,
biomedical similarity, and phenotype consistency. Vocabulary and output-budget constraints
prevent dense partial credit from rewarding fabricated or overlong differentials. Across eight
benchmarks, the complete RareDx system centered on Qwen3.5-9B reaches 38.34 macro Hit@10,
1.60 points above GPT-5.5 under the archived protocol; a disjoint validation-selection audit
retains a 6.80-point routing gain over Direct on held-out cases. The 27B system reaches
23.53/36.56/40.76 at Hit@1/5/10. Controlled
ablations show that retrieval is not uniformly helpful and that controlled routing is central
to the gain. These results indicate that structured medical knowledge can turn a compact model
into a competitive diagnostic ranker across heterogeneous long-tail settings in clinical practice.
\end{abstract}

\section{Introduction}
Rare diseases affect fewer than 1 in 2,000 individuals each, yet collectively affect more than
300 million people worldwide
\cite{valdez2016public,nguengang2020estimating,paul2013hope,jonker2024access,zhao2025agentic}.
Diagnosis commonly takes 4-5 years \cite{ghosh2025artificial}: phenotypes are incomplete,
heterogeneous, and shared with common disorders, increasing misdiagnosis risk
\cite{dong2020misdiagnosis}. Evidence is also dispersed across clinical records, genetic tests,
specialist knowledge, and disease resources. A useful system must retrieve the right evidence,
reject plausible distractors, and return a ranked differential of recognized disease entities.

Multimodal patient data and curated disease knowledge offer complementary diagnostic evidence
\cite{lee2022deep}. Foundation models, including LLMs \cite{thirunavukarasu2023large} and
vision-language models \cite{liu2025visual}, further broaden the information accessible to
medical AI systems
\cite{sarker2024natural,liu2024geneverse,tran2025multi,du2025accelerating,liu2025spemo,liu2025teampath}.
RareBench \cite{chen2024rarebench} and RareArena \cite{chen2026rarearena} evaluate general LLMs,
while agentic systems combine retrieval \cite{wang2025visualrag}, multi-agent communication
\cite{dhatterwal2023multi,chen2025enhancing}, and tools.
DeepRare \cite{zhao2025agentic} and Hygieia \cite{liu2026versatile}, for example, integrate
these components. Structured systems such as AI-MARRVEL \cite{mao2024ai}, LIRICAL
\cite{robinson2020interpretable}, and Exomiser \cite{smedley2015next} have demonstrated the
value of phenotype, genotype, and disease knowledge for prioritization.

However, two methodological limitations remain. First, agentic diagnostic systems are often
evaluated as monolithic pipelines, making it difficult to determine whether gains arise from
the backbone model, retrieval context, tool policy, repeated sampling, or output normalization.
Retrieval itself is not uniformly beneficial: irrelevant context can displace decisive
evidence even when the knowledge base contains the reference disease. Second, exact-match
rewards are too sparse for ranked rare-disease prediction. They assign the same failure signal
to a clinically related disorder and an arbitrary error, whereas unconstrained semantic
rewards may favor fabricated disease names or indiscriminate candidate enumeration. These
limitations are particularly consequential for compact open models, which require reliable
domain supervision to approach the diagnostic performance of closed-source LLMs in this
long-tail setting reliably.

We introduce \method{}, coupling RareDx-Harness for controlled knowledge use with RareDx-KGPO
for knowledge-graph-grounded post-training. The harness compares direct inference, static
retrieval, adaptive tools, structured phenotype-gene-disease reasoning, and aggregation under
one task and normalizer. RareDx-KGPO projects predictions into a medical graph and rewards
graded relevance, ontology proximity, semantic similarity, and phenotype consistency while
constraining invalid or overlong outputs. Across eight benchmarks, the complete 9B system
reaches 38.34 macro Hit@10, 1.60 points above GPT-5.5 under the archived protocol; a disjoint
selection audit retains a 6.80-point gain over Direct. The 27B system exceeds GPT-5.5 at all
three cutoffs. Ablations show that retrieval is not uniformly useful and that graph-aware,
controlled knowledge use drives the gain.

\section{Methods}

\subsection{Problem formulation}
Given record $x$, the model returns $\pi(x)=(d_1,\ldots,d_{10})$. The record may contain free
text, Human Phenotype Ontology (HPO) terms~\citep{kohler2021hpo}, or genetic findings. For the
accepted names and synonyms $\mathcal{G}(x)$, we report the normalized hit indicator
\begin{equation}
    \operatorname{Hit@}k(x)=\mathbbm{1}\!\left[\{d_1,\ldots,d_k\}\cap
    \mathcal{G}(x)\neq\varnothing\right],
\end{equation}
for $k\in\{1,5,10\}$ after deterministic disease canonicalization. Hit@10 is the primary
endpoint because the task requires a differential; Hit@1 isolates first-choice prioritization.

\subsection{RareDx-Harness: controlled diagnostic evaluation}
\begin{figure*}[t]
    \centering
    \includegraphics[width=0.98\textwidth]{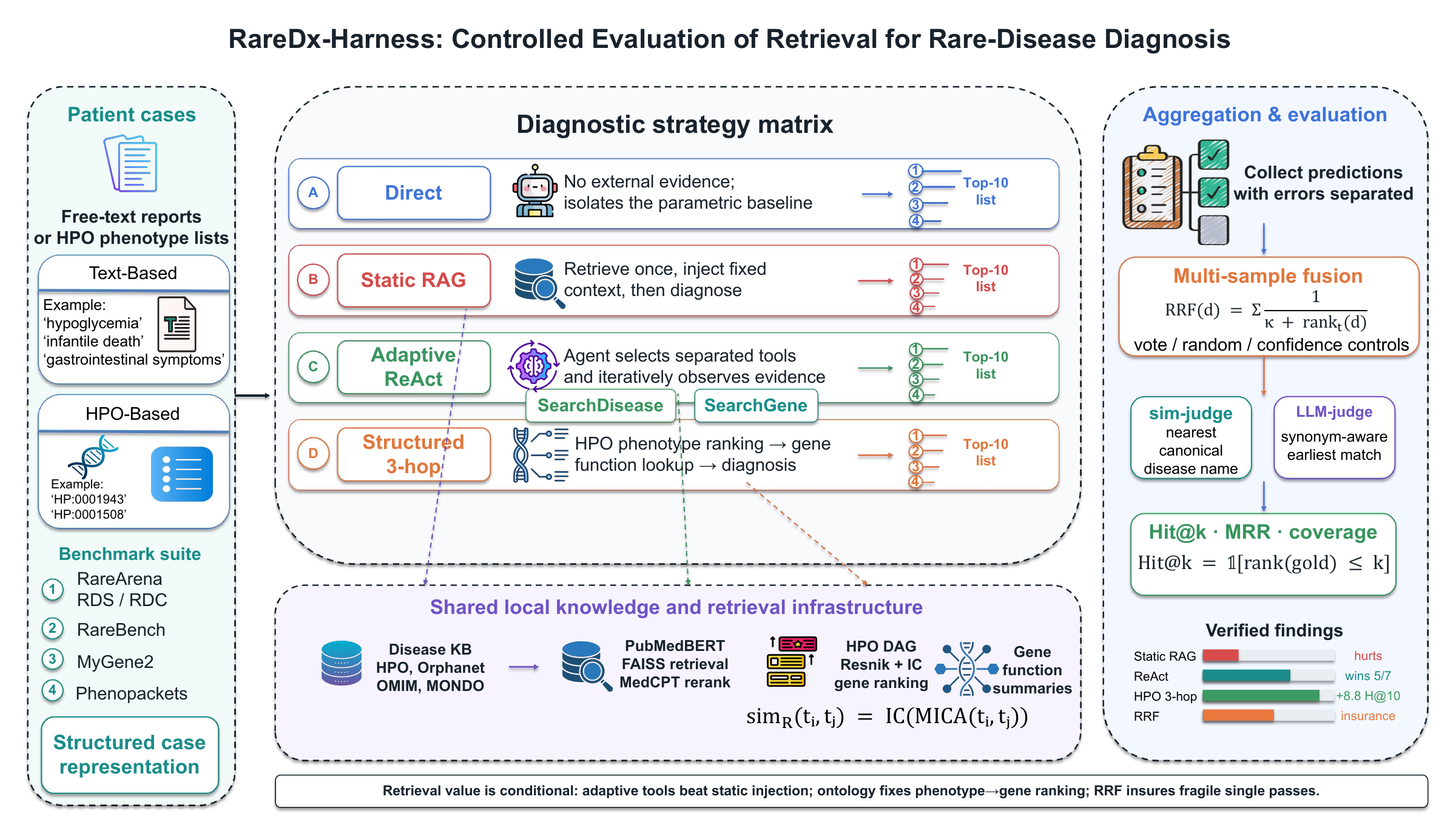}
    \caption{RareDx-Harness. Patient cases from heterogeneous benchmarks are normalized
    into a common ranked-diagnosis task. The controlled strategy matrix compares direct
    inference, one-shot static retrieval, adaptive ReAct tool use, and structured
    phenotype-to-gene-to-disease reasoning over the same local medical knowledge
    infrastructure. Canonicalization and rank aggregation are shared across strategies to
    preserve controlled comparison across all branches.}
    \label{fig:harness}
\end{figure*}

Figure~\ref{fig:harness} summarizes RareDx-Harness, which isolates parametric knowledge,
retrieval, tool policy, and stochastic aggregation. Every strategy consumes the same normalized
case and returns the same Top-10 schema, enabling paired comparison under one judge.

\paragraph{Diagnostic strategies.}
\textit{Direct} inference isolates the model's parametric knowledge. \textit{Static RAG}
retrieves once and prepends a fixed evidence block. \textit{Adaptive ReAct} exposes
separate disease and gene search tools, allowing the model to decide what to retrieve and
when to stop. \textit{Structured three-hop reasoning} first ranks genes from patient HPO
terms, obtains gene-function evidence, and then predicts diseases from the phenotype and
gene evidence jointly. For multi-sample inference, reciprocal-rank fusion (RRF) combines
lists without requiring calibrated generation probabilities.

\paragraph{Operational router.}
Routing is dataset-level rather than per patient: each model-benchmark pair chooses between
frozen Direct predictions and a deterministic 35B Diagnostic Audit that reranks the same list
without external tools. In the leakage-controlled audit, the route is selected on a fixed 20\%
development partition by Hit@10, then Hit@5 and Hit@1, and evaluated on the remaining 80\%.
Macro Hit@10 rises from 30.54 to 37.34 for 9B and from 29.83 to 38.13 for 27B. The complete
algorithm, prompts, candidate results, and ten-split stability analysis are in
Appendix~\ref{sec:router-audit}.

\paragraph{Shared evidence layer.}
All strategies share 27,554 records integrated from HPO~\citep{kohler2021hpo},
Orphanet~\citep{rath2012orphanet}, OMIM~\citep{amberger2019omim}, and Mondo
\citep{vasilevsky2026mondo}. Dense retrieval uses PubMedBERT, FAISS, and MedCPT
\citep{gu2021pubmedbert,douze2024faiss,jin2023medcpt}; structured reasoning uses the HPO graph,
Resnik similarity~\citep{resnik1995semantic}, and gene-function summaries. Because evidence and
output contracts are shared, performance differences reflect how each strategy uses the same
medical knowledge rather than unequal access or incompatible output formatting across methods.

\subsection{RareDx-KGPO: knowledge-graph-grounded policy optimization}
\begin{figure*}[t]
    \centering
    \includegraphics[width=0.98\textwidth]{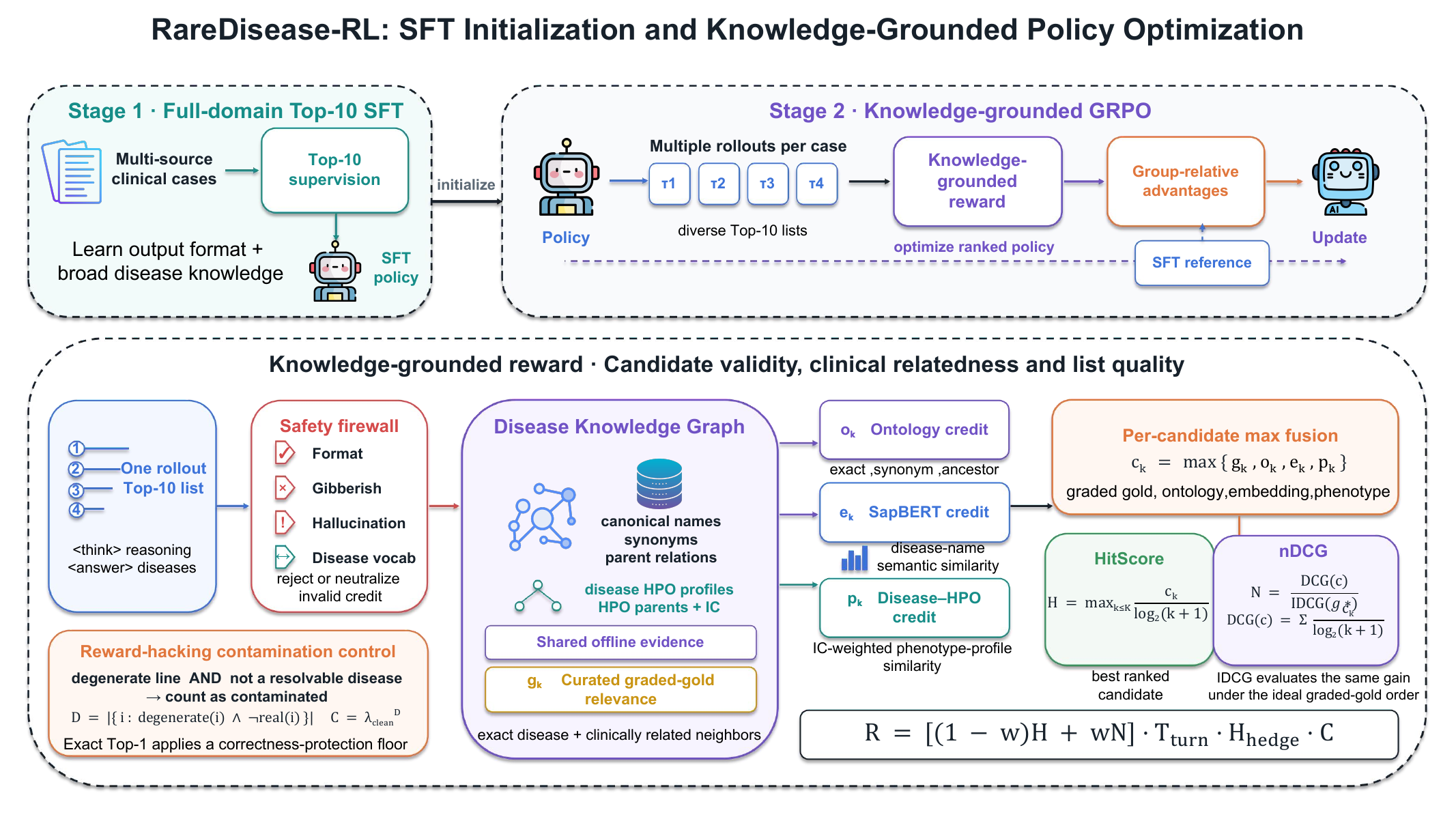}
    \caption{Two-stage post-training. Top-10 SFT establishes the output contract and broad
    diagnostic knowledge. RareDx-KGPO then optimizes ranked
    differential diagnoses using disease-graph, semantic, and phenotype evidence, with
    explicit safeguards against invalid names and reward hacking.}
    \label{fig:rl}
\end{figure*}

The two-stage recipe (Figure~\ref{fig:rl}) first teaches a valid Top-10 contract through SFT,
then samples multiple lists and computes group-relative advantages from graph-grounded clinical
rewards. The contribution is structured medical supervision and its safeguards, not a new
policy-gradient estimator. Zero-variance groups are resampled, policy-ratio clips are decoupled
\citep{shao2024deepseekmath,yu2025dapo}, and a reference KL term limits distribution shift.
Without resampling, uniformly wrong groups supply no relative signal; without asymmetric
clipping and reference regularization, dense rewards can shift the policy toward fluent but
invalid disease names and away from the diagnostic prior established during supervised fine-tuning.

\subsection{Projecting generations into a medical knowledge graph}
Each generated string is projected into a canonical space of disease aliases, ontology edges,
disease-HPO annotations, HPO ancestors, and information-content statistics. Exact matching is
followed by bounded edit-distance snapping; out-of-vocabulary strings retain their rank but
receive no dense credit. Every positive reward is therefore traceable to a known disease or
supported graph relation.

For candidate $y_k$ at rank $k$, four complementary evidence channels are evaluated.  Curated
graded-gold relevance $g_k$ assigns unit credit to the accepted diagnosis and its synonyms and
lower credit to verified clinically related diseases. Ontology credit is
\begin{equation}
    o_k=2^{-d_{\mathrm{onto}}(y_k,y^*)},
\end{equation}
when a graph path is available. Name-level credit $e_k$ is a thresholded, capped cosine
similarity between biomedical entity representations, preventing plausible names from matching
exact-diagnosis credit. Phenotype credit compares propagated profiles $P(y_k)$ and $P(y^*)$:
\begin{equation}
 p_k=\min\!\left(\tau_p,
 \frac{\sum_{h\in P(y_k)\cap P(y^*)}\operatorname{IC}(h)}
 {\sum_{h\in P(y_k)\cup P(y^*)}\operatorname{IC}(h)}\right).
 \label{eq:phenotype-credit}
\end{equation}
Information content emphasizes specific phenotypes; symmetric normalization avoids favoring
broad diseases with many nonspecific annotations.

The four channels encode different evidence. Curated labels are precise but sparse, ontology
paths are interpretable but incomplete, entity embeddings cover lexical variation but can
overvalue plausible phrasing, and phenotype overlap remains informative when disease edges are
missing. We therefore use maximum fusion
\begin{equation}
    c_k=\max\{g_k,o_k,e_k,p_k\}.
    \label{eq:candidate-credit}
\end{equation}
This retains the strongest supported relation without double-counting correlated evidence;
canonical projection prevents semantic similarity from rescuing fabricated names.

We retain a hit-oriented term
\begin{equation}
    H=\max_{k\le K}\frac{c_k}{\log_2(k+1)}
\end{equation}
and add a list-quality term
\begin{equation}
    N=\frac{\operatorname{DCG}@K}{\operatorname{IDCG}@K}
    =\frac{\sum_{k=1}^{K} g_k/\log_2(k+1)}
    {\sum_{k=1}^{K} g_k^{*}/\log_2(k+1)},
    \label{eq:ndcg}
\end{equation}
where $g_1^*,\ldots,g_K^*$ is the ideal graded ordering. $H$ prioritizes the best-supported
diagnosis, while $N$ separates lists with the same top candidate but different differential
quality. Using the same curated scale in its numerator and denominator prevents semantic
similarity from inflating nDCG and creates within-case variation. Without graded labels,
$w=0$ recovers $H$.

For a valid response, the trajectory reward is
\begin{equation}
 R_{\mathrm{valid}}=\big[(1-w)H+wN\big]
 T_{\mathrm{turn}}P_{\mathrm{hedge}}C_{\mathrm{clean}}.
 \label{eq:reward}
\end{equation}
Malformed outputs receive zero and globally degenerate outputs receive $-1$.
$P_{\mathrm{hedge}}$ penalizes overlong lists and $C_{\mathrm{clean}}=\rho^m$ penalizes $m$
degenerate, unresolvable lines, while valid long disease names and exact Top-1 answers are
protected. The single-turn experiments set $T_{\mathrm{turn}}=1$. Thresholds, transformations,
and inactive safeguards are specified in Appendix~\ref{sec:reward-details}.

\subsection{Evaluation setup}
We evaluate all cases from MyGene2, RAMEDIS, MME, HMS, LIRICAL, the RareArena RDS and RDC
tracks, and Phenopackets. Every method returns ten diseases scored by the same deterministic
normalizer; macro scores weight the eight sources equally. The principal model is Qwen3.5-9B
initialized by Top-10 SFT. Its archived run uses Equation~\ref{eq:reward} with $w=0$; the nDCG
extension is evaluated only on fixed held-out lists. Occamy-1.0~\citep{chen2026occamy} uses the
identical prompt, parser, and matcher. No training row shares a patient identifier or normalized
case text with evaluation; profile-level and seven-benchmark sensitivity audits preserve the
main comparison. Dataset sizes, provenance, overlap results, reward thresholds, and split limits
appear in Appendices~\ref{sec:reward-details} and~\ref{sec:data-audit}.

\section{Results}

\subsection{Overall diagnostic performance}
Table~\ref{tab:overall} gives 360 measurements from 15 systems, eight benchmarks, and three
cutoffs under one scoring protocol. RareDx-KGPO improves Qwen3.5-9B from
9.53/20.69/24.77 to 16.90/25.54/31.35 at Hit@1/5/10; the harness reaches
20.11/31.54/38.34. It exceeds GPT-5.5 by 1.60 points at Hit@10, is competitive at Hit@5,
and remains 1.90 points lower at Hit@1. RareDx-27B reaches
23.53/36.56/40.76 and exceeds GPT-5.5 by 1.52/4.97/4.02 points. These rows use the archived
dataset-level route and may invoke the 35B audit. Under disjoint development selection, 9B and
Appendix~\ref{sec:router-audit} reports the disjoint-selection result: 9B and 27B reach
37.34/38.13 Hit@10 versus 30.54/29.83 for Direct on matched held-out cases.

\begin{table*}[!t]
\centering
\caption{Complete diagnostic ranking results (\%) on eight benchmarks. The vertically
stacked panels report Hit@1, Hit@5, and Hit@10 within one unified table. Macro is the
unweighted mean across datasets. All outputs share one disease normalizer; bold marks
panel-best values.}
\label{tab:overall}
\fontsize{6.8}{7.4}\selectfont
\setlength{\tabcolsep}{3.0pt}
{\renewcommand{\arraystretch}{0.94}
\begin{tabular}{@{}lrrrrrrrrr@{}}
\toprule
\multicolumn{10}{c}{\textbf{(a) Hit@1: first-ranked diagnosis}} \\
\cmidrule(lr){1-10}
Model & MyGene2 & RAMEDIS & MME & HMS & LIRICAL & RDS & RDC & Pheno. & \textbf{Macro} \\
\midrule
\multicolumn{10}{l}{\textit{Closed-source LLMs}} \\
Claude Opus 4.7 & 9.6 & 21.1 & 12.5 & 22.2 & 23.4 & 12.4 & 15.2 & 25.6 & 17.8 \\
GLM-5.2 & 11.6 & 18.5 & 28.3 & 21.3 & 26.1 & 11.6 & 15.5 & 22.8 & 19.5 \\
GPT-5.5 & 20.5 & 21.5 & \textbf{29.5} & 23.8 & \textbf{27.2} & 10.7 & 12.1 & 30.8 & 22.0 \\
\midrule
\multicolumn{10}{l}{\textit{Open-weight baselines}} \\
MedGemma1.5-4B & 4.1 & 13.3 & 0.0 & 9.1 & 4.1 & 6.9 & 9.3 & 12.7 & 7.4 \\
Qwen3-8B & 5.5 & 12.8 & 0.0 & 18.2 & 9.7 & 4.8 & 6.7 & 5.0 & 7.8 \\
Qwen3.5-9B & 8.2 & 17.5 & 2.5 & 8.0 & 9.5 & 9.4 & 13.3 & 7.8 & 9.5 \\
Qwen3.6-27B & 8.2 & 20.4 & 7.5 & 20.4 & 12.7 & 12.5 & 15.6 & 12.0 & 13.7 \\
Qwen3.6-35B-A3B & 6.2 & \textbf{22.1} & 25.0 & 23.3 & 24.6 & 12.6 & 16.0 & 13.6 & 17.9 \\
Occamy-1.0 & 6.2 & 3.2 & 0.0 & 10.2 & 5.1 & 1.7 & 2.1 & 7.2 & 4.5 \\
Qwen3.6-35B-A3B (non-thinking) & 8.2 & 15.5 & 10.0 & 18.2 & 13.0 & 14.8 & 19.5 & 10.2 & 13.7 \\
Qwen3.8-27B & 6.2 & 20.8 & 7.5 & \textbf{27.3} & 13.0 & 12.8 & 16.3 & 11.2 & 14.4 \\
\midrule
\multicolumn{10}{l}{\textit{RareDx systems}} \\
\textbf{RareDx-9B, Direct (Ours)} & \textbf{21.9} & 10.4 & 5.0 & 12.5 & 21.4 & 10.4 & 17.6 & \textbf{36.0} & 16.9 \\
\textbf{RareDx-9B, Harness (Ours)} & \textbf{21.9} & 17.5 & 10.0 & 25.0 & 21.4 & 11.5 & 17.6 & \textbf{36.0} & 20.1 \\
\textbf{RareDx-27B, Direct (Ours)} & 15.8 & 9.9 & 5.0 & 12.5 & 21.4 & 16.0 & \textbf{42.0} & \textbf{36.0} & 19.8 \\
\textbf{RareDx-27B, Harness (Ours)} & 15.8 & 18.8 & 7.5 & \textbf{27.3} & 21.4 & \textbf{19.4} & \textbf{42.0} & \textbf{36.0} & \textbf{23.5} \\
\addlinespace[1.5pt]
\midrule
\multicolumn{10}{c}{\textbf{(b) Hit@5: diagnosis recovered within the top five}} \\
\cmidrule(lr){1-10}
Model & MyGene2 & RAMEDIS & MME & HMS & LIRICAL & RDS & RDC & Pheno. & \textbf{Macro} \\
\midrule
\multicolumn{10}{l}{\textit{Closed-source LLMs}} \\
Claude Opus 4.7 & 30.8 & 37.7 & 29.2 & 36.5 & \textbf{35.8} & 22.3 & 23.6 & 38.0 & 31.7 \\
GLM-5.2 & 22.6 & 33.8 & 32.6 & 37.7 & 35.0 & 21.7 & 24.9 & 30.6 & 29.9 \\
GPT-5.5 & \textbf{33.6} & 33.4 & 32.8 & 36.2 & 35.0 & 18.7 & 20.3 & \textbf{42.7} & 31.6 \\
\midrule
\multicolumn{10}{l}{\textit{Open-weight baselines}} \\
MedGemma1.5-4B & 17.1 & 21.0 & 0.0 & 20.5 & 8.6 & 15.5 & 18.2 & 30.9 & 16.5 \\
Qwen3-8B & 6.2 & 13.5 & 0.0 & 18.2 & 9.7 & 4.8 & 6.7 & 17.2 & 9.5 \\
Qwen3.5-9B & 20.5 & 37.3 & 7.5 & 23.9 & 16.5 & 19.6 & 23.4 & 16.8 & 20.7 \\
Qwen3.6-27B & 21.2 & 31.2 & 10.0 & \textbf{46.6} & 22.2 & 22.4 & 25.0 & 17.2 & 24.5 \\
Qwen3.6-35B-A3B & 16.4 & 36.0 & \textbf{33.3} & 36.7 & 35.1 & 20.8 & 22.9 & 20.2 & 27.7 \\
Occamy-1.0 & 18.5 & 8.5 & 2.5 & 14.8 & 10.8 & 7.6 & 10.5 & 17.4 & 11.3 \\
Qwen3.6-35B-A3B (non-thinking) & 12.3 & 37.2 & 10.0 & 38.6 & 20.8 & 25.1 & 28.8 & 22.0 & 24.4 \\
Qwen3.8-27B & 23.3 & 31.9 & 10.0 & 39.8 & 20.0 & 24.2 & 26.9 & 21.4 & 24.7 \\
\midrule
\multicolumn{10}{l}{\textit{RareDx systems}} \\
\textbf{RareDx-9B, Direct (Ours)} & 28.8 & 15.7 & 10.0 & 15.9 & 30.8 & 24.8 & 37.5 & 40.8 & 25.5 \\
\textbf{RareDx-9B, Harness (Ours)} & 28.8 & 37.7 & 10.0 & 39.8 & 30.3 & 27.1 & 37.8 & 40.8 & 31.5 \\
\textbf{RareDx-27B, Direct (Ours)} & 26.7 & 15.6 & 10.0 & 15.9 & 30.3 & 30.5 & \textbf{63.4} & 40.8 & 29.1 \\
\textbf{RareDx-27B, Harness (Ours)} & 26.7 & \textbf{40.7} & 10.0 & 42.0 & 30.3 & \textbf{38.6} & \textbf{63.4} & 40.8 & \textbf{36.6} \\
\addlinespace[1.5pt]
\midrule
\multicolumn{10}{c}{\textbf{(c) Hit@10: diagnosis recovered within the full differential}} \\
\cmidrule(lr){1-10}
Model & MyGene2 & RAMEDIS & MME & HMS & LIRICAL & RDS & RDC & Pheno. & \textbf{Macro} \\
\midrule
\multicolumn{10}{l}{\textit{Closed-source LLMs}} \\
Claude Opus 4.7 & 34.2 & 41.2 & \textbf{58.3} & 41.3 & \textbf{40.9} & 26.5 & 28.5 & 43.2 & 39.3 \\
GLM-5.2 & 30.8 & \textbf{47.6} & 39.1 & 41.0 & 39.0 & 26.7 & 29.3 & 34.4 & 36.0 \\
GPT-5.5 & 39.0 & 45.1 & 37.7 & 40.0 & 37.8 & 23.6 & 24.7 & \textbf{46.0} & 36.7 \\
\midrule
\multicolumn{10}{l}{\textit{Open-weight baselines}} \\
MedGemma1.5-4B & 23.3 & 22.8 & 2.5 & 23.9 & 11.1 & 19.9 & 22.2 & 38.2 & 20.5 \\
Qwen3-8B & 6.2 & 13.5 & 0.0 & 18.2 & 9.7 & 4.8 & 6.7 & 25.6 & 10.6 \\
Qwen3.5-9B & 25.3 & 39.6 & 10.0 & 33.0 & 19.7 & 23.3 & 26.7 & 20.6 & 24.8 \\
Qwen3.6-27B & 28.1 & 41.4 & 12.5 & \textbf{52.3} & 26.2 & 26.2 & 28.1 & 28.2 & 30.4 \\
Qwen3.6-35B-A3B & 25.3 & 41.9 & 41.7 & 40.0 & 40.2 & 24.7 & 26.4 & 23.6 & 33.0 \\
Occamy-1.0 & 26.0 & 10.7 & 2.5 & 15.9 & 14.6 & 11.2 & 16.7 & 19.6 & 14.7 \\
Qwen3.6-35B-A3B (non-thinking) & 17.8 & 41.3 & 10.0 & 47.7 & 24.9 & 29.2 & 32.7 & 27.6 & 28.9 \\
Qwen3.8-27B & 25.3 & 36.4 & 10.0 & 47.7 & 23.0 & 28.8 & 30.6 & 26.4 & 28.5 \\
\midrule
\multicolumn{10}{l}{\textit{RareDx systems}} \\
\textbf{RareDx-9B, Direct (Ours)} & \textbf{41.1} & 19.4 & 10.0 & 17.0 & 33.0 & 37.4 & 48.5 & 44.4 & 31.4 \\
\textbf{RareDx-9B, Harness (Ours)} & \textbf{41.1} & 42.3 & 10.0 & 50.0 & 31.9 & 37.8 & 49.2 & 44.4 & 38.3 \\
\textbf{RareDx-27B, Direct (Ours)} & 27.4 & 18.9 & 10.0 & 17.0 & 31.9 & 30.5 & \textbf{63.6} & 44.4 & 30.5 \\
\textbf{RareDx-27B, Harness (Ours)} & 27.4 & 45.7 & 12.5 & 51.1 & 32.1 & \textbf{49.3} & \textbf{63.6} & 44.4 & \textbf{40.8} \\
\bottomrule
\end{tabular}}
\vspace{-1mm}
\parbox{0.98\textwidth}{\textit{Notes:} Pheno. denotes Phenopackets testing samples.
Models are ordered consistently within each panel. RareDx model rows use Top-10 SFT with medical instruction tuning data followed
by RareDx-KGPO with rare-disease diagnosis data. Direct uses single-pass inference; Harness applies
the archived exploratory dataset-level route.}
\end{table*}

Gains concentrate on RAMEDIS and HMS for 9B and additionally on RDS for 27B; MME remains
difficult and 9B loses 1.1 Hit@10 points on LIRICAL. This heterogeneity motivates controlled
routing. Scale alone is insufficient: direct Qwen3.8-27B reaches 14.4/24.7/28.5, versus
23.5/36.6/40.8 for RareDx-27B. Occamy-1.0 reaches 4.46/11.32/14.65, showing that generic
co-work post-training does not directly transfer to this specialization.
The three cutoffs also separate candidate coverage from first-choice accuracy. Harness gains
grow with $k$, indicating that controlled evidence primarily improves the clinically useful
differential before consistently resolving its top rank. This is preferable to reporting a
single cutoff that could conceal shallow prioritization gains or indiscriminate expansion of
the differential without improving diagnostic prioritization.

\subsection{RL algorithm comparison}
GRPO uses sequence rewards and group-relative normalization; DAPO adds dynamic sampling and
decoupled clipping; OPD distills dense teacher distributions on student trajectories
\citep{agarwal2024onpolicy}. RareDx-KGPO instead applies Equation~\ref{eq:reward} without a
rollout teacher; Appendix Table~\ref{tab:rl-algorithms} summarizes the supervision and update
distinctions.

\begin{figure*}[t]
    \centering
    \includegraphics[width=0.96\textwidth]{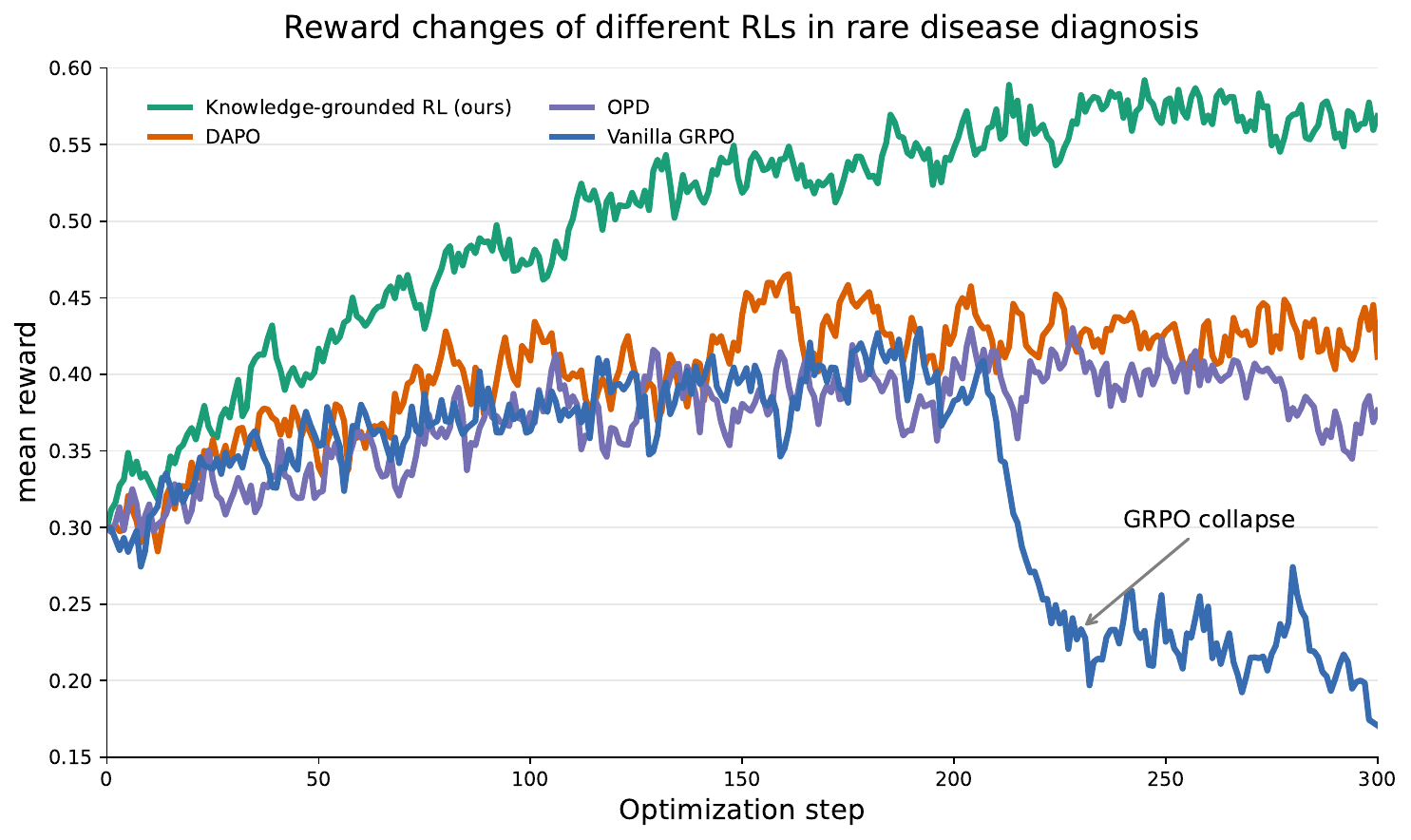}
    \caption{Reward dynamics for rare-disease post-training from a shared initialization.
    Vanilla GRPO improves initially but collapses after approximately 210 steps, consistent
    with reward hacking through indiscriminate disease-name generation. DAPO and OPD avoid
    catastrophic collapse but converge to lower rewards. RareDx-KGPO (Ours) supports sustained
    improvement and remains stable at the final checkpoint without late-stage collapse.}
    \label{fig:reward-trajectories}
\end{figure*}

From the shared initialization in Figure~\ref{fig:reward-trajectories}, GRPO reaches 0.40 but
collapses after step 210 as indiscriminate disease-name generation exploits semantic partial
credit. DAPO and OPD remain near 0.40 but plateau or decline. RareDx-KGPO reaches approximately
0.58 without late collapse because vocabulary projection, capped partial credit, graph and
phenotype evidence, and anti-hedging penalties close these reward shortcuts.
Inspection of late GRPO rollouts confirms the corresponding behavioral change: outputs become
weakly conditioned on the patient and enumerate disease names that obtain incidental semantic
credit. RareDx-KGPO instead keeps positive credit tied to canonical entities and graph-supported
relations, so higher reward remains aligned with a bounded, patient-specific differential
rather than a generic disease list.

\subsection{Reward-component ablation and adversarial audit}
On 120 held-out diagnoses, the phenotype channel gives the nearest non-reference HPO neighbor
0.316 mean reward versus 0.008 without it, while exact diagnoses remain at 1.0. The vocabulary
gate rejects every constructed pseudo-disease; removing it rewards 82.5\%. With rank 1 fixed,
the list-aware term ($w=0.3$) scores coherent, reversed, and unrelated lists at
1.000/0.997/0.911, whereas hit-only reward ties them. Constructions, intervals, and sensitivity
curves are in Appendix~\ref{sec:reward-audit}.

\subsection{Harness comparison and component ablation}
\begin{figure*}[t]
    \centering
    \includegraphics[width=0.88\textwidth]{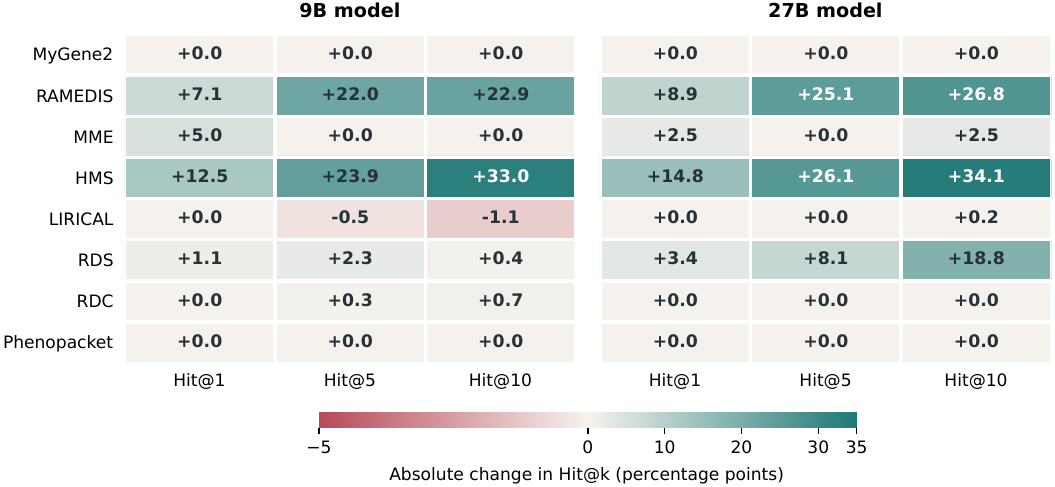}
    \caption{Per-dataset effect of applying the archived exploratory RareDx-Harness route
    to the same knowledge-grounded checkpoint. Cells show the absolute change in Hit@1,
    Hit@5, and Hit@10. Gains concentrate on RAMEDIS and HMS at both scales, with an additional
    large 27B gain on RDS. Near-zero cells correspond to retaining Direct when the audit
    provided no measured benefit; the small LIRICAL regression illustrates that
    routing is not error-free.}
    \label{fig:harness-delta}
\end{figure*}

Figure~\ref{fig:harness-delta} shows gains of 3.21/6.00/6.99 points at Hit@1/5/10 for 9B and
3.71/7.41/10.30 for 27B. Larger gains at deeper cutoffs indicate improved differential
coverage, but the effect is dataset-dependent, supporting case-aware rather than uniform
evidence acquisition.

Static RAG underperforms Direct, showing that evidence volume is not evidence quality.
Table~\ref{tab:harness-components} shows that ReAct widens coverage but slightly reduces
Hit@1. On Phenopackets, HPO-Resnik raises gene recall@8 from 15.3\% to 69.4\% and Hit@10 from
27.5\% to 36.3\%, whereas dense gene retrieval adds less than one point. Structured retrieval
can nevertheless introduce hard negatives elsewhere, explaining the need for routing.
RRF should therefore be interpreted as variance reduction across stochastic lists, not as a
remedy for systematically misleading evidence. The router is useful precisely because no
acquisition strategy dominates across the heterogeneous records and evidence regimes represented
by these benchmarks in our evaluation.

\begin{table}[!ht]
\centering
\caption{Component ablations with fixed backbone and judge. Panel A uses Qwen3.6-Flash and
macro averages over seven benchmarks. Panel B uses Qwen3.6-35B-A3B on Phenopackets.}
\label{tab:harness-components}
% \resizebox{\columnwidth}{!}{%
\begin{tabular}{llrrr}
\toprule
Panel & Configuration & Hit@1 & Hit@5 & Hit@10 \\
\midrule
A & Direct & 17.03 & 26.63 & 31.27 \\
A & Adaptive ReAct & 16.49 & 27.51 & 33.53 \\
\midrule
B & Direct & 17.2 & 23.7 & 27.5 \\
B & 3-hop, dense gene retrieval & 16.2 & 23.6 & 28.4 \\
B & 3-hop, HPO-Resnik & \textbf{22.2} & \textbf{31.7} & \textbf{36.3} \\
\bottomrule
\end{tabular}
\end{table}

An adaptive-depth audit further finds a non-monotonic association between tool calls and
Hit@10: one and two calls score 57.96 and 59.31, versus 43.32 at three calls. Because depth is
policy-selected, this is descriptive rather than causal. Appendix~\ref{sec:retrieval-depth}
reports the complete strata, confidence intervals, and depth-conditioned diagnostic results.

\begin{figure*}[t]
    \centering
    \includegraphics[width=0.70\textwidth]{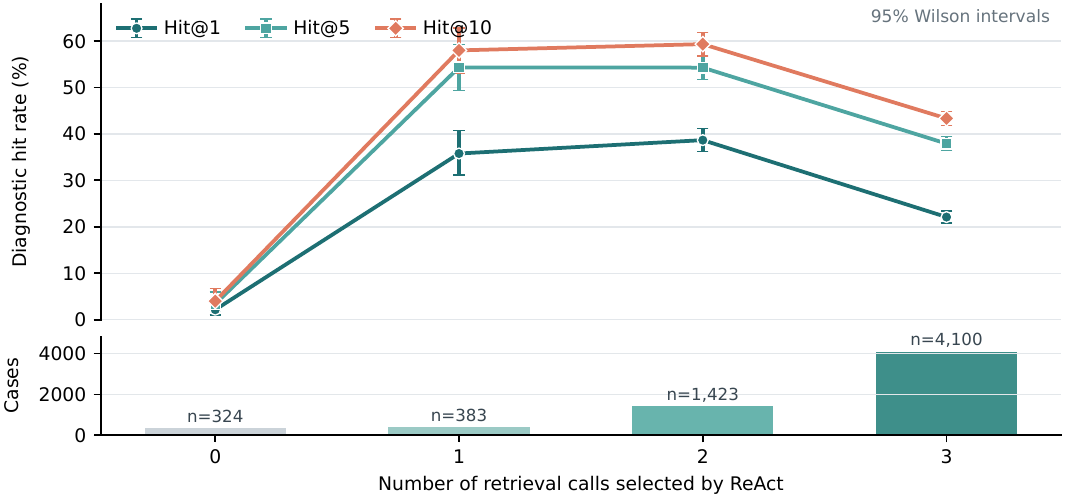}
    \caption{Observed accuracy by retrieval depth for 6,230 ReAct trajectories, with 95\%
    Wilson intervals \cite{wilson1942confidence} and stratum sizes for binomial testing. Depth is policy-selected, so the comparison describes
    behavior rather than estimating the causal effect of retrieval depth.}
    \label{fig:react-depth}
\end{figure*}
\FloatBarrier

\subsection{Case study: from recognition to a usable differential}
We audit the first 50 RAMEDIS records as a fixed slice using matched prompts, greedy decoding,
token budget, parser, and normalizer. Direct 27B exceeds direct 9B at Hit@10, but the 9B Harness
reverses the comparison and reaches 14/38/46 at Hit@1/5/10 (Table~\ref{tab:case-slice}). This
paired audit diagnoses system behavior; it does not replace the eight-benchmark result.

\begin{table}[!ht]
\centering
\caption{Paired audit on the first 50 RAMEDIS records. Values are percentages under the disease
matcher and parsing contract used throughout.}
\label{tab:case-slice}
% \resizebox{\columnwidth}{!}{%
\begin{tabular}{lrrr}
\toprule
System & Hit@1 & Hit@5 & Hit@10 \\
\midrule
Qwen3.8-27B, direct & 2 & 12 & 28 \\
Ours 9B, direct & 4 & 12 & 16 \\
Ours 9B, full Harness & \textbf{14} & \textbf{38} & \textbf{46} \\
\bottomrule
\end{tabular}
\end{table}

The 9B system recovers 16 diagnoses missed by direct 27B and loses seven, a net gain of nine
Hit@10 cases; 13 recoveries enter the Top-5 rather than merely extending the tail. Under the
same 1,024-token budget, direct 27B averages 462 words and never completes the required answer
block, whereas the post-trained 9B model averages 91 words and completes all 50. Appendix
\ref{sec:case-details} provides parsing details and a clinical example in which the Harness
converts broad fatty-acid-oxidation recognition into the correct MCADD ranking.
In that case, hypoglycemia, gastrointestinal symptoms, elevated transaminases, abnormal
carnitine, and infantile death support a fatty-acid-oxidation disorder. The larger direct model
recognizes the mechanism but does not produce a usable ranked answer before truncation. RareDx
ranks MCADD first and organizes the remaining differential around related oxidation and
carnitine-transport disorders, illustrating how the output contract and controlled evidence
turn broad recognition into an actionable ranking.

\begin{figure*}[t]
    \centering
    \includegraphics[width=0.72\textwidth]{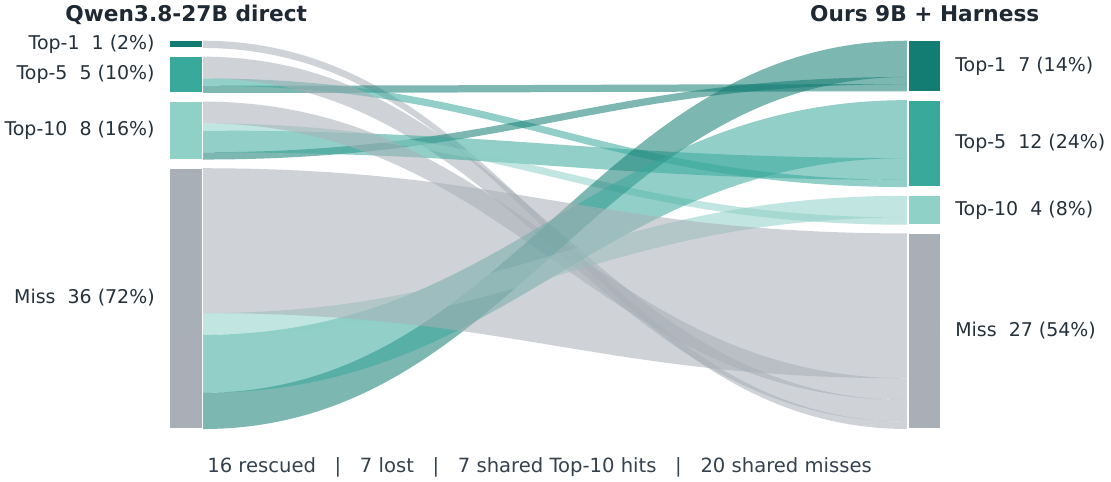}
    \caption{Paired rank-category transitions on the fixed 50-case RAMEDIS audit. Each ribbon
    follows one patient from direct Qwen3.8-27B to RareDx-9B with the full Harness. RareDx
    recovers 16 diagnoses missed by the larger model, 13 of which enter the Top-5.}
    \label{fig:case-transition}
\end{figure*}

\subsection{Efficiency analysis}
On the same eight-A800 setup and 6,249 records, direct 9B and 27B generation takes 137.2 and
340.0 seconds, or 45.5 versus 18.4 cases/s, giving 9B a 2.48-fold throughput advantage.
These timings cover direct batched generation and exclude loading, scoring, retrieval, and tool
latency. Structured retrieval adds local computation without another model call; ReAct adds
variable model-mediated turns. Appendix
Table~\ref{tab:efficiency} reports the complete computational profile of every strategy.

\section{Discussion}
Rare-disease diagnosis benefits from structured knowledge, but retrieval is not monotonic:
static context and high-recall gene candidates can introduce persuasive hard negatives, making
routing part of the method. Conversely, ontology, phenotype, and curated relevance provide
dense supervision while preserving entity validity and rank priority. Together these choices
let RareDx-9B exceed GPT-5.5 at macro Hit@10 under the archived protocol and let RareDx-27B
exceed it at every cutoff. The paired RAMEDIS audit confirms a system effect: direct 9B trails
direct 27B, whereas the full 9B system reverses the comparison.
More broadly, the results distinguish knowledge availability from knowledge use: the same
store may help or distract depending on acquisition and case representation. RareDx makes this
distinction measurable through a common output contract and learnable through entity-grounded
rewards. Its 9B advantage therefore reflects controlled evidence acquisition, normalization,
and ranking rather than a claim of greater parametric medical knowledge. Reliable interfaces
to structured resources can rival backbone scale in long-tail tasks, especially at Hit@5 and
Hit@10, where coherent coverage matters alongside first-choice confidence.

Limitations include dataset heterogeneity, the small MME split, and automatic rather than
clinical adjudication. Archived routes reflect exploratory development; disjoint-selection
results are the controlled estimate. Routed 9B results may invoke a 35B audit model, and direct
throughput excludes loading, retrieval, and tool latency. Reward trajectories characterize
optimization but do not replace matched downstream evaluation of every checkpoint. Finally,
benchmark accuracy does not establish clinical safety; prospective studies must assess
calibration, harmful omission, evidence faithfulness, and robustness across care settings.
Future work should learn patient-level routing exclusively from development data, evaluate
calibrated abstention, and measure whether retrieved evidence supports the final rank rather
than merely correlating with it. Prospective studies should also report subgroup performance
and clinician revision rates, since a plausible but misplaced rare diagnosis may impose costs
that Hit@$k$ alone cannot capture.

\FloatBarrier
\clearpage
\subsubsection*{Use of Large Language Models}
We used a large-language-model-based coding assistant during the development of this work.
The assistant supported code review and refactoring, figure generation, and improvements to
the clarity and presentation of the manuscript. The authors specified the research questions,
designed the methods and experiments, interpreted the results, and reviewed all
assistant-supported changes. All reported results were obtained from the described
experiments, and all citations were checked against identifiable source publications. The
authors take full responsibility for the manuscript and did not use language models to
fabricate experimental evidence, citations, or supporting claims.

\subsubsection*{Ethics Statement}
RareDx is developed for academic research on computational rare-disease diagnosis and is not
intended to provide clinical advice or replace evaluation by qualified clinicians. Model
outputs may be incomplete, incorrect, biased, or unsupported, particularly for underrepresented
conditions and patient populations. Accordingly, the system should not be used for diagnosis,
treatment selection, or other clinical decisions without independent expert review and
appropriate validation. Users are responsible for complying with applicable requirements for
patient privacy, data governance, informed consent, and responsible disclosure. Derivative
models and materially modified system configurations should be released under distinct version
identifiers so that their provenance and relationship to the evaluated artifacts remain
traceable. We encourage explicit documentation of model, knowledge-base, and evaluation
versions in any subsequent research use.

\subsubsection*{Reproducibility Statement}
We archive the task definitions, verifier configurations, reference annotations, per-run
predictions, evaluation summaries, model matrix, and ablation data underlying the reported
tables and figures. The numerical results and plots are generated programmatically from these
artifacts using fixed evaluation scripts. Training and implementation details, including hardware, decoding settings,
optimization parameters, and checkpoint-selection criteria, are provided in the appendix. Our codes and model weights will be released after peer review.

\bibliography{iclr2027_conference}
\bibliographystyle{iclr2027_conference}

\newpage 

\appendix
\section{Benchmark composition}
The macro score gives each of the eight reported benchmark sources equal weight despite
their different sample counts. RareBench is evaluated on all questions, with RAMEDIS,
MME, HMS, and LIRICAL reported separately to expose source-level variation.
Table~\ref{tab:dataset-size} records the composition used throughout the main results.
\begin{table}[htbp]
\centering
\caption{Number of test records contributing to the macro average. RareBench coverage
includes all questions, reported separately for RAMEDIS, MME, HMS, and LIRICAL.}
\label{tab:dataset-size}
\begin{tabular}{lrlr}
\toprule
Dataset & Cases & Dataset & Cases \\
\midrule
MyGene2 & 146 & RAMEDIS & 624 \\
MME & 40 & HMS & 88 \\
LIRICAL & 370 & RDS & 1,803 \\
RDC & 678 & Phenopackets & 500 \\
\midrule
\multicolumn{3}{r}{Total} & 4,249 \\
\bottomrule
\end{tabular}
\end{table}

\section{Harness router specification and leakage-controlled audit}
\label{sec:router-audit}
The phrase ``selected Harness strategy'' in Table~\ref{tab:overall} refers to a
\emph{dataset-level} router, not a per-case learned gate.  Its two executable candidates are
(i) \textsc{Direct}, the frozen RareDx Top-10 response, and (ii) \textsc{Audit}, a deterministic
diagnostic audit by a locally served Qwen3.6-35B-A3B model.  The auditor receives the original
patient evidence and the Direct ranked list, is instructed to retain supported entries and repair
unsupported or missing diagnoses, and returns ten canonical disease names.  It uses greedy
decoding (temperature 0), a 384-token output cap, disabled thinking, and no retrieval or external
tools.  If parsing yields fewer than two audit candidates, the Direct list is retained; otherwise
the audit list is deduplicated and any unfilled tail is copied from Direct.  The four inference
strategies compared in the main Harness study are research configurations; the final router only
chooses between the two archived candidates above.

To make route selection independent of the evaluation portion, we add a strict audit over frozen
outputs.  For every model-dataset pair, a SHA-256 hash of \texttt{seed|dataset|case-id} assigns
approximately 20\% of cases to development and the remainder to test (seed 20260919).  Let
$m_s^D=(H@10,H@5,H@1)$ be strategy $s$'s development tuple.  The router is
\begin{equation}
 s_D^*=\arg\max_{s\in\mathcal S_D}^{\mathrm{lex}}m_s^D,
 \qquad \mathcal S_D\subseteq\{\mathrm{Direct},\mathrm{Audit}\},
 \label{eq:router}
\end{equation}
where the lexicographic comparison prioritizes Hit@10, then Hit@5 and Hit@1; an exact tie selects
\textsc{Direct}, the cheaper strategy.  The decision is made once per model and benchmark source,
never per patient.  Audit is included in $\mathcal S_D$ only when a complete frozen audit output
exists for that source.  No test score enters Equation~\ref{eq:router}.

\begin{table}[htbp]
\centering
\caption{Validation-only route audit. Each metric cell is Hit@1/5/10 (\%). D and A denote
Direct and Diagnostic Audit. A dash means that a complete archived Audit candidate was not
available, so the candidate set contained Direct only.}
\label{tab:router-strict}
\scriptsize
\setlength{\tabcolsep}{3.0pt}
\begin{tabular}{llrrlll ll}
\toprule
Model & Dataset & $n_{dev}$ & $n_{test}$ & D, dev & A, dev & Pick & D, test & A, test \\
\midrule
9B & MyGene2 & 33 & 113 & 24.2/27.3/39.4 & 12.1/18.2/27.3 & D & 21.2/29.2/41.6 & 14.2/29.2/32.7 \\
9B & RAMEDIS & 113 & 511 & 11.5/18.6/21.2 & 16.8/40.7/47.8 & A & 9.6/14.9/18.4 & 17.6/37.0/41.1 \\
9B & MME & 9 & 31 & 0.0/11.1/11.1 & 0.0/0.0/0.0 & D & 6.5/9.7/9.7 & 12.9/12.9/12.9 \\
9B & HMS & 22 & 66 & 22.7/27.3/27.3 & 27.3/45.5/59.1 & A & 9.1/12.1/13.6 & 24.2/37.9/47.0 \\
9B & LIRICAL & 62 & 308 & 17.7/27.4/29.0 & 16.1/27.4/30.7 & A & 22.1/30.8/32.5 & 19.2/27.0/30.8 \\
9B & RDS & 342 & 1,461 & 11.7/27.2/38.3 & -- & D & 10.1/24.2/37.2 & -- \\
9B & RDC & 127 & 551 & 18.1/36.2/48.8 & -- & D & 17.4/37.8/48.5 & -- \\
9B & Phenopackets & 103 & 397 & 42.7/47.6/50.5 & -- & D & 34.3/39.0/42.8 & -- \\
\midrule
27B & MyGene2 & 33 & 113 & 9.1/24.2/24.2 & 6.1/21.2/21.2 & D & 17.7/27.4/28.3 & 14.2/26.6/29.2 \\
27B & RAMEDIS & 113 & 511 & 14.2/21.2/21.2 & 17.7/46.0/50.4 & A & 10.8/24.3/24.3 & 19.0/39.5/44.6 \\
27B & MME & 9 & 31 & 0.0/0.0/0.0 & 0.0/0.0/0.0 & D & 3.2/3.2/3.2 & 9.7/12.9/16.1 \\
27B & HMS & 22 & 66 & 36.4/36.4/36.4 & 31.8/54.6/68.2 & A & 22.7/27.3/27.3 & 25.8/37.9/45.5 \\
27B & LIRICAL & 62 & 308 & 16.1/19.4/19.4 & 16.1/29.0/29.0 & A & 14.9/21.1/21.1 & 16.2/27.0/29.9 \\
27B & RDS & 342 & 1,461 & 15.8/30.4/30.4 & 19.6/38.6/48.0 & A & 16.0/30.5/30.5 & 19.4/38.6/49.6 \\
27B & RDC & 127 & 551 & 40.2/58.3/59.1 & 26.8/47.2/57.5 & D & 42.5/64.6/64.6 & 29.8/48.3/64.4 \\
27B & Phenopackets & 103 & 397 & 43.7/44.7/44.7 & -- & D & 30.0/39.3/39.3 & -- \\
\bottomrule
\end{tabular}
\end{table}

Across the eight test partitions, the validation-only router obtains 18.81/30.21/37.34 for 9B
and 21.71/34.69/38.13 for 27B.  We also repeat the complete selection protocol for ten fixed hash
seeds.  Audit is selected in all ten splits for 9B RAMEDIS and HMS and for
27B RAMEDIS, LIRICAL, and RDS.  Selection is less stable for the small MME set and for MyGene2,
and it varies for 9B LIRICAL and 27B RDC.  This sensitivity is why the strict audit reports every
candidate rather than only the selected route.

The headline routing in Table~\ref{tab:overall} was produced earlier during exploratory
development using benchmark-level results; it was not selected under the strict split protocol
above.  Consequently, Table~\ref{tab:router-strict}, rather than the headline routed row, is the
appropriate estimate when validation-only route selection is required.  The frozen Direct rows
remain unaffected by this distinction.

\section{Exact reward implementation}
\label{sec:reward-details}
Let $y_k$ be the disease string at rank $k\leq 10$.  Normalization removes punctuation and
parenthetical aliases and maps $y_k$ to the nearest canonical vocabulary entry only when normalized
Levenshtein distance is at most 0.20.  An unmapped string receives no dense medical credit unless
it is explicitly present in the curated graded-gold set.  For a mapped candidate, the four reward
components shown in Figure~\ref{fig:rl} are
\begin{align}
 g_k &= \max_{r\in\mathcal C}\gamma(r)\,
        \mathbb{1}[\operatorname{norm}(y_k)=r],\\
 o_k &= \max_{g\in G}2^{-d_{\mathcal O}(y_k,g)},\\
 e_k &= \max_{g\in G} f(\cos(z_{y_k},z_g)),\\
 p_k &= \min\{0.40,\operatorname{WJaccard}_{IC}(P(y_k),P(y^*))\},\\
 c_k &= \max\{g_k,o_k,e_k,p_k\},
\end{align}
where $\mathcal C$ is the curated set containing the accepted diagnosis and clinically related
neighbors, $\gamma(r)\in(0,1]$ is their graded relevance, $G$ is the accepted diagnosis set,
$d_{\mathcal O}$ is shortest-path distance in the disease ontology, and $P(y^*)$ is the explicit
reference HPO set when provided or otherwise the
knowledge-graph profile of the reference disease.  The symmetric phenotype overlap is weighted
by HPO information content.  The embedding transform is
\begin{equation}
f(s)=\begin{cases}
0, & s\leq0.75,\\
0.30(s-0.75)/0.05, & 0.75<s<0.80,\\
0.30+0.05(s-0.80)/0.10, & 0.80\leq s<0.90,\\
0.35, & s\geq0.90.
\end{cases}
\label{eq:embedding-transform}
\end{equation}
Thus exact or synonymous ontology matches receive 1, while one- and two-edge relations receive
0.5 and 0.25.  The rank-sensitive base score is
$H=\max_{k\leq10}c_k/\log_2(k+1)$.

The anti-hacking transform is applied after medical scoring.  A candidate line is contaminated
when it is degenerate and cannot be resolved to a real disease; degeneracy is triggered by more
than 12 words, nonword-character mass above 0.5, or token repetition of at least 0.5.  If $m$ lines
are contaminated, the cleanliness multiplier is $2^{-m}$.  Outputs with more than ten raw
candidates receive a 0.9 budget multiplier.  To protect a verified exact rank-1 answer from a
residual false-positive contamination heuristic, the deployed configuration floors its cleanliness
multiplier at 1.  A valid anchor requires ontology distance at most one
or embedding cosine at least 0.80.  A response is globally invalid when at least half of its lines
are degenerate or fewer than half are resolvable; a globally invalid response without an anchor
receives $-1$, while a missing or empty \texttt{<answer>} block receives 0.  For the reported
checkpoint, define $q=1$ for an exact ontology match at rank 1 and $q=2^{-m}$ otherwise.  In the
general formulation illustrated in Figure~\ref{fig:rl},
\begin{equation}
R=\big[(1-w)H+wN\big]T_{\mathrm{turn}}H_{\mathrm{hedge}}C_{\mathrm{clean}}.
\end{equation}
For the reported single-turn run, $w=0$, $T_{\mathrm{turn}}=1$,
$H_{\mathrm{hedge}}=0.9^{\mathbbm{1}[L>10]}$, and $C_{\mathrm{clean}}=q$, giving
\begin{equation}
R=H\,q\,0.9^{\mathbbm{1}[L>10]},
\label{eq:implemented-reward}
\end{equation}
subject to the malformed and globally invalid overrides above.  The list-level nDCG extension is
implemented in the analysis code but its weight is zero for the reported training run; it is used
only in the fixed-output audit in Section~\ref{sec:reward-audit}.  This distinction prevents the
audit-only term from being attributed to the trained checkpoint.

\section{Training provenance and overlap audit}
\label{sec:data-audit}
Table~\ref{tab:training-data} records the exact artifacts referenced by the 9B launch scripts.
SFT converts each source answer into an ordered Top-10 target using seed 13, placing the accepted
diagnosis first and filling alternatives from teacher predictions, KG phenotype neighbors,
ontology neighbors, then lexical or random fallbacks when graph coverage is insufficient.  The RL
mixture totals 61,918 prompts.  The RareArena article-backed portion contains recorded publication
dates from 2003-08-18 through 2024-06-15 for 22,403 of 42,977 rows; the remaining artifacts do not
store a reliable source-publication date, so no broader temporal cutoff is claimed.  Train and
validation are separate materialized files.  The archived preparation code does not retain the
original randomization manifest, which prevents reconstructing a stronger chronological split
claim after the fact.

\begin{table}[htbp]
\centering
\caption{Training and validation artifacts used by the reported 9B pipeline.}
\label{tab:training-data}
\small
\begin{tabular}{llrl}
\toprule
Stage & Component & Rows & Content \\
\midrule
SFT & train & 51,368 & RareArena cases with constructed Top-10 targets \\
SFT & validation & 932 & held-out Phenopacket-format prompts \\
RL & RareArena & 42,977 & article-backed case reports and test results \\
RL & Phenopacket & 8,392 & phenotype/gene disease prompts \\
RL & gene-phenotype hard set & 3,928 & synthetic hard cases \\
RL & simulated HPO subset & 2,714 & synthetic incomplete-phenotype cases \\
RL & sparse/atypical set & 3,907 & synthetic sparse or atypical cases \\
RL & validation & 932 & Phenopacket-format prompts \\
\bottomrule
\end{tabular}
\end{table}

We performed a read-only overlap audit against all eight evaluation sources.  Three signatures are
checked independently: normalized patient/case identifier, SHA-1 of normalized case text, and exact
or near-exact disease-plus-phenotype set (HPO Jaccard at least 0.90).  There are no identifier or
case-text matches in any SFT or RL training artifact.  SFT has no phenotype-signature match.  RL
training has one exact disease-plus-phenotype signature shared with LIRICAL (1 of 61,918 prompts),
without a shared identifier or case text.  The RL validation file shares 290 disease-plus-phenotype
profiles with the 500-case Phenopackets benchmark, again with no identifier or case-text match.
These are profile-level matches and do not establish patient identity, but they can make
Phenopackets-based checkpoint selection optimistic.  As a conservative sensitivity analysis,
removing Phenopackets from the macro average leaves RareDx-9B at 37.47 Hit@10 versus 35.41 for
GPT-5.5, increasing rather than reversing the reported margin.

The retrieval collection is intentionally not entity-disjoint from evaluation: it contains 27,554
disease-level documents built from HPO, Orphanet, OMIM, and MONDO, plus an optional 2,299 gene
documents.  These records contain disease names, definitions, synonyms, and phenotype/gene
associations but not benchmark patient reports.  Entity and phenotype overlap is therefore the
intended retrieval signal, whereas patient-text overlap is not.  Source snapshot dates were not
embedded in the archived indices, so we do not claim a temporal-disjointness guarantee for the
retrieval corpus.

\section{Proprietary baseline protocol}
\label{sec:proprietary-protocol}
All proprietary rows use the same task template: a clinical-geneticist instruction, the patient
phenotypes or case report (and test results for RDC), and a requirement to return exactly ten
specific formal disease names in descending order, one per line, without explanation.  The system
message is \texttt{You are a helpful assistant.}  No retrieval, function call, web search, or other
tool is enabled, and the intended protocol makes one generation per case with no self-consistency
aggregation.

\begin{table}[htbp]
\centering
\caption{Recorded proprietary-baseline invocation. ``Provider default'' means the argument is
absent from the archived request rather than assigned an inferred value.}
\label{tab:proprietary-config}
\small
\begin{tabular}{p{0.18\textwidth}p{0.20\textwidth}p{0.23\textwidth}p{0.27\textwidth}}
\toprule
Reported row & Request alias & Interface & Decoding and tools \\
\midrule
GPT-5.5 & \texttt{gpt-5.5} & OpenAI-compatible proxy, \texttt{xiaoai.plus/v1}
& temperature, top-$p$, seed, and output cap: provider default; tools: none \\
Claude Opus 4.7 & \texttt{claude-opus-4-7} & OpenAI-compatible proxy,
\texttt{xiaoai.plus/v1} & temperature, top-$p$, seed, and output cap: provider default; tools: none \\
GLM-5.2 & \texttt{glm-5.2} & Alibaba MaaS compatible API for phenotype sets;
DashScope Generation for RDS/RDC & decoding and output cap: provider default; tools: none \\
\bottomrule
\end{tabular}
\end{table}

The notebooks retain aliases but not immutable provider snapshot identifiers, response headers,
run timestamps, or provider-side default values.  Therefore the exact serving revisions and
numeric decoding defaults cannot be recovered and are not inferred here.  This is a limitation of
the original baseline collection.  Future releases will persist request JSON, resolved model
revision, response metadata, per-case attempt count, and raw output for every proprietary call.

\section{Post-training configurations}
Table~\ref{tab:rl-algorithms} separates the optimization rule, supervision signal, and need
for an online teacher. This distinction is important because our contribution is the medical
graph reward rather than a new group-normalization update.
\begin{table}[htbp]
\centering
\caption{Optimization and supervision differences in the RL comparison.}
\label{tab:rl-algorithms}
\begin{tabular}{llll}
\toprule
Method & Update & Signal & Teacher \\
\midrule
GRPO & Group relative & Task reward & No \\
DAPO & Dynamic groups, asymmetric clip & Task reward & No \\
OPD & On-policy distillation & Token distributions & Yes \\
RareDx-KGPO (Ours) & Group relative & Medical graph reward & No \\
\bottomrule
\end{tabular}
\end{table}

\section{Case-study details}
\label{sec:case-details}
The case study uses row 12 of the frozen RAMEDIS evaluation split. Its phenotype list is
\{death in infancy, hypoglycemia, vomiting, diarrhea, elevated hepatic transaminases,
abnormal circulating carnitine concentration\}. Both models receive the same system
instruction and ranked-differential template, use greedy decoding, and have a maximum
generation length of 1,024 tokens. The complete 9B Harness list is: medium-chain acyl-CoA
dehydrogenase deficiency; carnitine palmitoyltransferase II deficiency; very-long-chain
acyl-CoA dehydrogenase deficiency; multiple acyl-CoA dehydrogenase deficiency;
carnitine-acylcarnitine translocase deficiency; primary systemic carnitine deficiency;
mitochondrial trifunctional protein deficiency; short-chain acyl-CoA dehydrogenase
deficiency; glutaric acidemia type II; and propionic acidemia.

For the 50-case audit, we take rows 0-49 before inspecting outputs. The direct 27B outputs
are newly generated, while direct 9B and Harness outputs come from the frozen evaluation
artifacts used in the main experiment. All three are rescored with the same parser and name
matcher. The audit contains only RAMEDIS cases and is deliberately reported separately from
the eight-dataset macro average.

\begin{table}[h]
\centering
\caption{Observed outputs for RAMEDIS case 12.}
\label{tab:case}
\small
\begin{tabular}{p{0.16\columnwidth}p{0.76\columnwidth}}
\toprule
Model & Observed output \\
\midrule
Qwen3.8-27B & Restates phenotype categories and surveys metabolic mechanisms, but does not
produce a disease-ranked answer before truncation. MCADD is not recovered. \\
Ours 9B + Harness & Ranks MCADD first, followed by carnitine palmitoyltransferase II
deficiency, very-long-chain acyl-CoA dehydrogenase deficiency, multiple acyl-CoA
dehydrogenase deficiency, and carnitine-acylcarnitine translocase deficiency. \\
\bottomrule
\end{tabular}
\end{table}

\begin{table}[h]
\centering
\caption{Direct-generation behavior on the 50-case audit. Answer-block compliance requires
both opening and closing tags. Output words are counted after whitespace tokenization and are
reported only as a model-agnostic verbosity diagnostic.}
\label{tab:case-format}
\begin{tabular}{lrrr}
\toprule
Direct model & Complete block & Mean words & Median words \\
\midrule
Qwen3.8-27B & 0/50 & 462.2 & 451.0 \\
Ours 9B & 50/50 & 91.3 & 90.5 \\
\bottomrule
\end{tabular}
\end{table}

\section{Reward-audit construction}
\label{sec:reward-audit}
\begin{figure*}[t]
    \centering
    \includegraphics[width=0.88\textwidth]{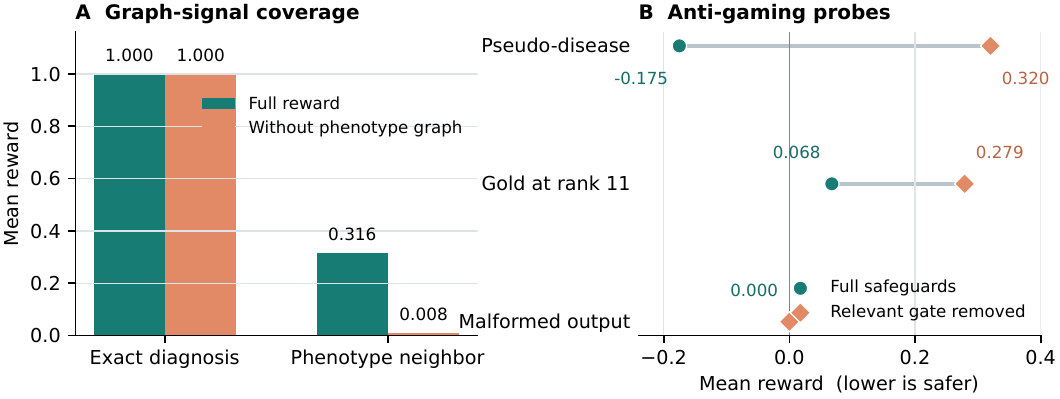}
    \caption{Controlled reward audit on 120 held-out diagnoses. Panel A removes the
    disease-phenotype graph signal while holding outputs fixed. Panel B removes the
    vocabulary gate for pseudo-diseases or the Top-10 budget for a delayed exact match.}
    \label{fig:reward-ablation}
\end{figure*}

We select the first 120 validation diseases that resolve to the frozen disease graph and have
at least one distinct disease sharing an HPO annotation. For the phenotype-neighbor probe, we
choose the non-reference disease maximizing set Jaccard similarity between the two disease HPO
profiles. The pseudo-disease probe prepends and appends diagnostic modifiers to the normalized
reference name, creating a medically styled string that is absent from the canonical
vocabulary. The over-budget probe places ten real distractor diseases before the reference.
All prompts retain their original validation metadata. Biomedical name embeddings are disabled
for both the full and ablated configurations in this audit, while the same graph, disease
profiles, parser, and normalization thresholds are retained.

\begin{table}[h]
\centering
\caption{Numerical reward audit. ``Ablated'' removes the phenotype-graph channel, vocabulary
gate, or Top-10 budget for the corresponding probe. Positive is the percentage of constructed
outputs receiving reward greater than zero.}
\label{tab:reward-audit}
\begin{tabular}{lrrrr}
\toprule
Probe & Full & Ablated & Positive, full & Positive, ablated \\
\midrule
Exact diagnosis & 1.000 & 1.000 & 100.0 & 100.0 \\
Phenotype neighbor & 0.316 & 0.008 & 100.0 & 0.8 \\
Pseudo-disease & -0.175 & 0.320 & 0.0 & 82.5 \\
Gold at rank 11 & 0.068 & 0.279 & 100.0 & 100.0 \\
Malformed output & 0.000 & 0.000 & 0.0 & 0.0 \\
\bottomrule
\end{tabular}
\end{table}

\subsection{List-quality discrimination audit}
For each eligible validation disease, we rank other graph diseases by Jaccard overlap of their
HPO profiles and retain the four strongest distinct neighbors. Their graded relevance is the
observed overlap capped at 0.6; the reference receives 1.0. The coherent list places these
neighbors in descending relevance after the rank-1 reference, the reverse list flips only
their order, and the unrelated list replaces them with deterministic real-disease distractors.
All configurations use the same parser, vocabulary gate, Top-10 budget, disease graph, and
phenotype channel; embeddings are disabled to avoid cross-signal substitution. Confidence
intervals use 2,000 case-level bootstrap resamples with a fixed seed. At $w=0.3$, the
unrelated-list mean is 0.911 (95\% CI 0.907-0.914), while the coherent-list mean is 1.000.
This experiment also exposed and corrected a normalization error: the DCG numerator must use
the same curated relevance set as IDCG, rather than arbitrary dense similarity. The headline
$w=0$ checkpoints are unaffected by this correction.

\begin{figure*}[t]
    \centering
    \includegraphics[width=0.90\textwidth]{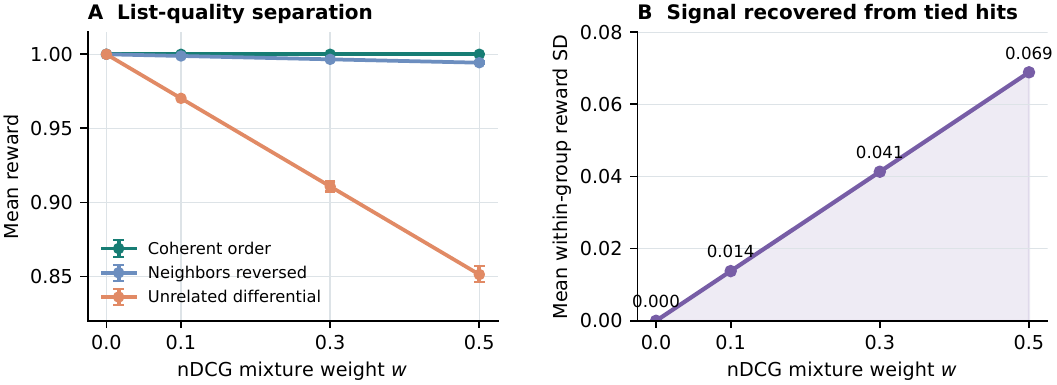}
    \caption{List-quality audit on 120 held-out diagnoses. Every list contains the exact
    diagnosis at rank 1. Panel A sweeps the nDCG weight with case-level bootstrap 95\%
    intervals. Panel B reports within-case reward variation across coherent, reversed, and
    unrelated differentials.}
    \label{fig:ndcg-discrimination}
\end{figure*}

\section{Adaptive retrieval-depth audit}
\label{sec:retrieval-depth}
Table~\ref{tab:react-depth} provides the numerical values underlying
Figure~\ref{fig:react-depth}. The frozen Qwen3.7-Plus judge returns the rank of the accepted
reference diagnosis in each generated list, or zero when it is absent. We pool the four
evaluation sources only for this behavioral analysis and retain every archived trajectory.
Wilson intervals in the figure are computed independently within each depth stratum.

\begin{table}[h]
\centering
\caption{ReAct diagnostic performance stratified by the number of retrieval calls selected
by the policy. Values are percentages.}
\label{tab:react-depth}
\begin{tabular}{rrrrr}
\toprule
Calls & Cases & Hit@1 & Hit@5 & Hit@10 \\
\midrule
0 & 324 & 2.16 & 3.40 & 4.01 \\
1 & 383 & 35.77 & \textbf{54.31} & 57.96 \\
2 & 1,423 & \textbf{38.65} & 54.25 & \textbf{59.31} \\
3 & 4,100 & 22.10 & 37.90 & 43.32 \\
\bottomrule
\end{tabular}
\end{table}

\begin{table}[t]
\centering
\caption{Computational profile. $M$ is the number of sampled lists and $T$ the number of
ReAct turns.}
\label{tab:efficiency}
\begin{tabular}{lccc}
\toprule
Strategy & LLM generations & Local retrieval & External API \\
\midrule
Direct & 1 & None & No \\
Static RAG & 1 & Once & No \\
Adaptive ReAct & $T$ & Adaptive & No \\
Structured 3-hop & 1 & HPO + gene lookup & No \\
RRF fusion & $M$ & Strategy-dependent & No \\
\bottomrule
\end{tabular}
\end{table}

\section{Efficiency measurement details}

The archived 9B and 27B runs use eight data-parallel workers, tensor parallelism of one,
greedy decoding, a 4,096-token context limit, and the same 6,249-record generation suite.
The suite includes the main test records and auxiliary validation splits, so its size differs
from Table~\ref{tab:dataset-size}. Per-shard
generation times for 9B range from 128.4 to 137.2 seconds; the corresponding 27B range is
321.8 to 340.0 seconds. Node-level throughput is computed as the total number of cases divided
by the maximum shard time, since evaluation completes when the slowest worker finishes. Model
initialization, result merging, disease normalization, and metric computation are excluded.

\section{Training and implementation details}
The 9B RL run uses two nodes with eight A800 80GB GPUs per node. The global training batch
and generation batch are both 512, with 16 rollouts per prompt and a PPO minibatch size of
128. Prompt and response limits are 1,024 and 768 tokens. The actor learning rate is
$5\times10^{-7}$; the reference-policy KL coefficient is 0.15. The lower and upper policy
ratio clips are 0.20 and 0.28, respectively. Training rollouts use temperature 1.0, while
validation uses 0.7. Groups with zero reward variance are resampled, training runs for at
most 300 steps, and checkpoints are written every ten steps. The reported checkpoint is
selected on the held-out validation evaluation rather than training reward alone.
\section{Supplementary model scale ablation}
\label{app:sampling-budget}
We isolate the effect of additional sampled diagnosis lists in a supplementary
128-case development-set experiment. This experiment uses fixed round-60,
rank-64 LoRA checkpoints of Qwen3.8-27B and Qwen3.6-35B-A3B. These are separate
checkpoints and a separate evaluation protocol from the main benchmark tables;
the following values should not be compared directly with those tables.
The cases are drawn from the 932-case development pool and are disjoint from an
earlier 128-case pilot, but are not a new held-out test set. The validation-overlap
limitations discussed above still apply. Scoring uses normalized exact matches
to the gold diagnosis and its supplied aliases, without vocabulary snapping or
an LLM judge.

For each case, we retain the first diagnosis from a greedy list and fill the
remaining nine positions by reciprocal-rank voting over $k$ additional lists.
Each unique name contributes $1/r$ at its first position $r$ in each sampled list;
ties are broken lexicographically. No gold labels are used in aggregation.
Additional lists use temperature 1.0, top-$p=0.95$, top-$k=50$, and a 1,024-token
output limit, without retrieval or tools. All settings reuse the same recorded
outputs: $k=1,2,4$ use fixed prefixes of the eight samples. These intermediate
budgets were examined post hoc; the $k=8$ aggregation rule was fixed before this
expanded evaluation. Top-1 is preserved by construction, at 28.91\% for 27B and
28.12\% for 35B.

\begin{table}[t]
\centering
\small
\begin{tabular}{r r r r r}
\toprule
 & \multicolumn{2}{c}{27B} & \multicolumn{2}{c}{35B} \\
Extra lists $k$ & Hit@10 (\%) & Output tokens & Hit@10 (\%) & Output tokens \\
\midrule
0 & 34.38 & 173 & 32.81 & 179 \\
1 & 36.72 & 346 & 32.81 & 354 \\
2 & 39.06 & 519 & 36.72 & 529 \\
4 & 39.84 & 864 & 42.19 & 877 \\
8 & 46.09 & 1,556 & 42.19 & 1,575 \\
\bottomrule
\end{tabular}
\caption{Sampling-budget ablation on 128 development cases with fixed model
weights. Output tokens are mean totals per case across all $k+1$ generations;
they exclude input tokens, retrieval, and runtime overhead.}
\label{tab:sampling-budget-dev}
\end{table}

At $k=8$, Hit@10 increases by 11.72 percentage points for 27B and 9.38 points
for 35B relative to greedy decoding, at approximately nine times the generated-token
budget. For 35B, $k=4$ and $k=8$ have the same aggregate Hit@10 on this subset,
although their correct cases differ; this does not establish general saturation.
The experiment supports a coverage--generation-budget trade-off for this
aggregation rule, not an improvement in model weights or first-choice accuracy.
It also does not isolate the causal benefit of voting from that of extra sampling.

\end{document}